\documentclass[letterpaper, 10 pt, conference]{IEEEtran}  

\IEEEoverridecommandlockouts                              

\usepackage[utf8]{inputenc}

\usepackage[hidelinks]{hyperref}
\usepackage[style=ieee,hyperref,natbib=true,backend=bibtex,firstinits,doi=false,%
     mincitenames=1,maxcitenames=2,maxbibnames=99,sorting=none,terseinits=false,hyperref=true]{biblatex}
\bibliography{references}
\renewbibmacro*{bbx:savehash}{}
\defbibheading{bibliography}[\bibname]{\section*{References}}

\usepackage{url}
\usepackage{graphicx}
\usepackage[usenames,dvipsnames]{xcolor}
\usepackage[draft,inline,nomargin]{fixme}
\fxsetup{theme=color}

\usepackage{amsmath}
\usepackage{amssymb}
\usepackage{bbm}
\usepackage{textcomp}
\usepackage{siunitx}

\usepackage{booktabs}
\usepackage{threeparttable}
\usepackage{multirow}

\usepackage{mathtools}

\usepackage[capitalize]{cleveref}

\usepackage{tikz}
\usepackage{tikz-3dplot}
\usepackage[normalem]{ulem}   
\usetikzlibrary{arrows}
\usetikzlibrary{arrows.meta}
\usetikzlibrary{positioning,calc}
\usetikzlibrary{decorations.pathreplacing}
\usetikzlibrary{decorations.markings}
\usetikzlibrary{fit}
\usetikzlibrary{shapes.callouts}
\usetikzlibrary{shapes.geometric}
\usetikzlibrary{shapes.misc}
\usetikzlibrary{matrix}
\usetikzlibrary{spy}
\tikzstyle{line}=[draw]
\usepackage{setspace}
\usepackage{csquotes}

\usepackage{pgfplots}
\pgfplotsset{compat=1.9}
\pgfplotsset{every tick label/.append style={font=\scriptsize}}
\usepgfplotslibrary{groupplots}
\usepgfplotslibrary{units}
\usepgfplotslibrary{colorbrewer}
\usepackage{pgfplotstable}

\usepackage{balance}
\usepackage{xfrac}

\usepackage{blindtext}

\usepackage{ifthen}
\usepackage{tabularx}
\usepackage{makecell}

\usepackage[export]{adjustbox}

\IfFileExists{graphicscache.sty}{\usepackage{graphicscache}}

\DeclareMathOperator*{\argmin}{arg\,min}

\usepackage{mathtools}
\usepackage{placeins}
\usepackage{dblfloatfix}

\usepackage{array}
\newcolumntype{P}[1]{>{\centering\arraybackslash}p{#1}}
\newcolumntype{M}[1]{>{\centering\arraybackslash}m{#1}}

\DeclarePairedDelimiterX\Basics[1](){ #1}

\newcommand{\probP}{\text{I\kern-0.15em P}}

\newcommand{\PNP}{P\textit{n}P}

\title{\LARGE \bf
  Efficient Multi-Object Pose Estimation using Multi-Resolution Deformable Attention and Query Aggregation
\thanks{$^{*}$ Equal contribution}
}

\author{
\IEEEauthorblockN{Arul Selvam Periyasamy$^{*}$}
\IEEEauthorblockA{\textit{Autonomous Intelligent Systems} \\
\textit{University of Bonn}\\
Bonn, Germany \\
Email: {\tt periyasa@ais.uni-bonn.de}}
\and
\IEEEauthorblockN{Vladimir Tsaturyan$^{*}$}
\IEEEauthorblockA{\textit{Autonomous Intelligent Systems} \\
\textit{University of Bonn}\\
Bonn, Germany \\
Email: {\tt vladimir@gmail.com}}
\and
\IEEEauthorblockN{Sven Behnke}
\IEEEauthorblockA{\textit{Autonomous Intelligent Systems} \\
\textit{University of Bonn}\\
Bonn, Germany \\
Email: {\tt behnke@cs.uni-bonn.de}}
}

\begin{document}

\maketitle

\begin{abstract}
Object pose estimation is a long-standing problem in computer vision.
Recently, attention-based vision transformer models have achieved state-of-the-art results in many computer vision applications.
Exploiting the permutation-invariant nature of the attention mechanism, a family of vision transformer models formulate multi-object pose estimation as a set prediction problem.
However, existing vision transformer models for multi-object pose estimation rely exclusively on the attention mechanism.
Convolutional neural networks, on the other hand, hard-wire various inductive biases into their architecture. 
In this paper, we investigate incorporating inductive biases in vision transformer models for multi-object pose estimation, which facilitates learning long-range dependencies while circumventing the costly global attention.
In particular, we use multi-resolution deformable attention, where the attention operation is performed only between a few deformed reference points.
Furthermore, we propose a query aggregation mechanism that enables increasing the number of object queries without increasing the computational complexity.
We evaluate the proposed model on the challenging YCB-Video dataset and report state-of-the-art results. 

\end{abstract}


\section{Introduction}
Object pose estimation is the task of predicting the position and the orientation of objects with respect to the sensor coordinate frame.
It plays a vital role in many autonomous robotic systems and augmented and virtual reality applications.
Occlusion, object reflectance properties, and lighting conditions increase the complexity of the task.
Despite the recent deep-learning-driven progress, object pose estimation remains challenging.
RGB-D methods, in general, perform better than the RGB-only methods.
The depth information facilitates easier learning of the geometric features compared to the RGB input.
However, transparent and reflecting objects present significant challenges for \mbox{RGB-D} cameras.
The resolution and frame rate of the \mbox{RGB-D} sensors are limited, compared to RGB cameras.
Additionally, in large-scale real-world deployment, RGB-D sensors need calibration, which is time and resource-consuming. 
Motivated by these limitations of the RGB-D sensors, we focus on RGB methods in this paper.

\begin{figure}
  \centering
  \resizebox{.8\linewidth}{!}{\input{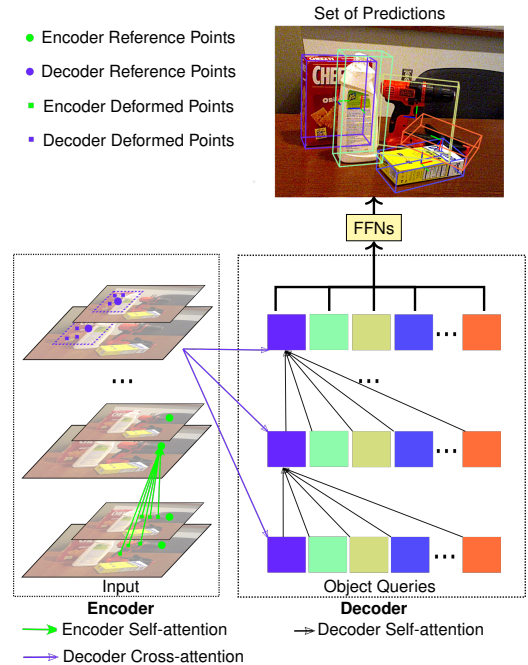}}
    \vspace*{-3mm}
    \caption{Multi-resolution deformable attention model.
    Our architecture utilizes an attention-based encoder-decoder design.
    In the encoder, image patch features of multiple resolutions are extracted using linear projection, and the self-attention mechanism is used to generate encoder embeddings.
    In the decoder, cross-attention is performed between the encoder embeddings and learned embeddings known as \emph{object queries} to generate object embeddings.
    The object queries are initialized randomly at the beginning of training and learned jointly with the training objective.
    During inference, the object queries remain fixed.
    In contrast to standard attention, in which attention is computed between all image patch features, in deformable attention, 
    the attention operation is performed only between the deformed reference points.
    From the object embeddings, object predictions are generated using feed-forward neural networks (FFNs) in parallel.
    Object pose predictions generated by our model are visualized using 3D bounding boxes.}
   \label{fig:model}
   \vspace{-2mm}
\end{figure}

In the last decade, convolutional neural networks (CNNs) have become the standard machine learning tool for solving many computer vision tasks.
The success of CNNs is largely attributed to the inductive biases incorporated into their architectural design~\citep{lecun1995convolutional, behnke2003hierarchical, CohenS17, schulz2012deep}.
Lately, transformer-based architectures have found great success in many computer vision tasks.
The core component of the transformer model is the multi-head attention mechanism that allows learning dependencies in the input data over a long range.
\citet{carion2020end} introduced DETR, an end-to-end differentiable architecture for object detection combining a CNN model for feature extraction and
a transformer-based encoder-decoder model for generating a set of object predictions. Inspired by their work, we formulate multi-object pose estimation as a set prediction problem.
However, instead of relying completely on the computationally costly global attention mechanism, to facilitate efficient architectures, we investigate CNN-like inductive biases in our design.

\noindent In this work, we present the following contributions:
\begin{itemize}
\item an efficient multi-resolution deformable attention model for multi-object pose estimation,
\item a query aggregation mechanism to increase the number of object queries without increasing the computational complexity, and
\item a thorough evaluation on the \mbox{YCB-Video} dataset~\citep{xiang2017posecnn} with state-of-the-art results.
\end{itemize}

\section{Related Work}
\subsection{Object Pose Estimation}
\label{sec:related_pose}
The prominent deep learning methods can be classified as direct regression~\citep{xiang2017posecnn,arash2021gcpr,Wang_2021_GDRN},
keypoint-based~\citep{rad2017bb8,peng2019pvnet,hu2020single}, 
and refinement-based~\citep{li2018deepim,labbe2020,periyasamy2019refining}.
The direct regression methods formulate pose estimation as a regression task. 
Given the image crop consisting of the target object, the direct regression methods predict the
6D pose parameters. In contrast, the keypoint-based methods estimate the pixel coordinates of
3D keypoints in the image and with known 2D-3D correspondence, the 6D pose is recovered employing
the perspective-\textit{n}-point (\PNP) algorithm.
The refinement-based methods are orthogonal to both direct regression and keypoint-based methods in terms of the approach.
They formulate pose estimation as a problem of pose refinement using the render-and-compare framework.
With the notable exception of \citet{capellen2019convposecnn} and \citet{hu2020single}, most of the 
CNN architectures for object pose estimation decouple pose estimation from object detection
and follow a two-stage approach in which the 2D bounding boxes are estimated in the first stage;
and in the second stage, only the crop containing the target object is processed to estimate the 6D pose.
To realize an end-to-end differentiable architecture for pose estimation, these methods rely on
modules like non-maximum suppression (NMS), region of interest (ROI) pooling, anchor box proposal~\citep{ren2015faster, hosang2017learning,redmon2017yolo9000},
differentiable implementations of the Hough transform~\citep{xiang2017posecnn,capellen2019convposecnn}, 
and random sample consensus (RANSAC)~\citep{brachmann2017dsac}.

\subsection{Vision Transformers}

\citet{vaswani2017attention} introduced the Transformer architecture with a multi-head attention mechanism to model
long-range dependencies in natural language processing and achieved significant improvements on a variety of tasks.
The success of the transformer architecture inspired many approaches that incorporate attention mechanisms to solve computer vision tasks
either by supplementing or by replacing CNN modules.
\citet{dosovitskiy2020image} introduced Vision Transformer (ViT), an architecture without any CNN modules.
While ViT achieved impressive results using a simple architecture, the computational cost of attention was higher than CNN modules. 
Subsequent methods showed that incorporating CNN-like inductive biases into the vision transformer architecture reduced computational costs and improved performance across a wide range of tasks.
\citet{liu2021Swin} introduced the Swin Transformer model, which limits self-attention to local non-overlapping shifted windows and 
uses cross-attention for cross-window connections.
They also incorporated hierarchical processing into their architecture.
\citet{li2021focal} introduced focal self-attention, an efficient attention mechanism in which the pixels in the closest surrounding tokens are attended at a fine granularity but the pixels far away are processed at a coarse granularity.
DETR~\cite{carion2020end} formulated multi-object detection as a set prediction problem and proposed a transformer-based encoder-decoder architecture
for set prediction.
\citet{zhu2021deformable} proposed deformable attention to reduce computational cost in DETR-like architectures.
\citet{arash2021gcpr} extended DETR for multi-object 6D pose regression.
\citet{amini2022yolopose, periyasamy2023yolopose} introduced YOLOPose, a DETR-like architecture for keypoint-based multi-object pose estimation. 
In contrast to the multi-stage pose estimation models discussed in \cref{sec:related_pose},
they realized a single-stage multi-object pose estimation without using  NMS, ROI, or anchor box proposal modules.

\section{Method}
\label{sec:method}

In this section, we introduce the multi-object pose estimation as a set prediction task formulation and 
briefly describe the existing YOLOPose model~\citep{amini2022yolopose} that we use as the baseline.
Later, we discuss the improvements to the baseline model incorporating various inductive biases. 

Multi-head attention (MHA) introduced by \citet{vaswani2017attention} performs scaled dot-product attention between the
query-key pairs.
Let $q \in \Omega_q$ be a query element with feature $z_q \in \mathbf{R}^{d}$ and
$k \in \Omega_k$ be a key element with feature $x_k \in \mathbf{R}^{d}$, where $d$ is the dimension of the features and
$\Omega_{q/k}$ are the sets of query and key elements, respectively.
Then the multi-head attention feature is computed as: 
\begin{multline}
  \label{eqn:mha}
\text {MHA}(z_q, x) = \sum_{m=1}^{M} W_m \left[  \sum_{k\in\Omega_k} A_{mqk} \cdot W^{'}_mx_k \right],
\end{multline}

where $m \in M$ represents the attention head, $W^{'}_m \in \mathbf{R}^{d_v \times d }$  and $W_m \in \mathbf{R}^{d \times d_v}$ 
are learnable projection parameters, \mbox{$d_v = d/M$}, and $A$ represents the normalized attention weight. 
MHA is the core component of the transformer architecture.


\begin{figure}
  \centering
\resizebox{.5\linewidth}{!}{\input{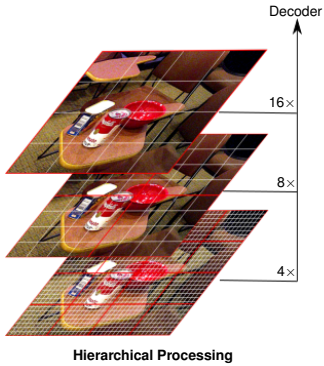}}
\vspace*{-2mm}
    \caption{Hierarchical processing of image features. White grids represent the image patches and red grids represent the self-attention windows.
    Layers are grouped into blocks.
    The size of the self-attention window and the number of feature maps remain fixed for all the layers inside each block
    but increase through the hierarchy.
    Restricting attention to a local window helps reduce the computational complexity.}
    \label{fig:swin_hierarchical}
\end{figure}

\begin{figure}
  \centering
  \resizebox{.8\linewidth}{!}{\input{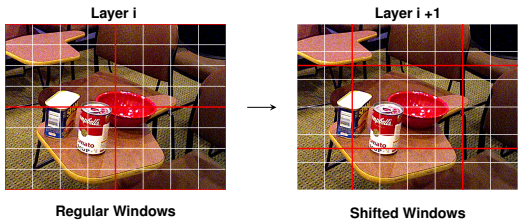}}
    \vspace*{-3mm}
    \caption{ 
    Shifting window local attention mechanism.
    Self-attention is restricted to the local window shown in red grids.
    The windows are shifted between the layers, enabling global interaction within each block.}
    \label{fig:swin_windows}
\end{figure}

Formulating pose estimation as a set prediction problem enables multi-object pose estimation in which objects are detected and their 6D pose are estimated in one single forward pass.
Given the set of object predictions $\hat{\mathcal{Y}}$ of cardinality $N$ and ground-truth set $\mathcal{Y}$ padded with \text{\O} class, we search for the optimal permutation
$\hat{\sigma}$ among the possible permutations $\sigma \in \mathfrak{S}_{N}$
defined by the matching cost $\mathcal{L}_{match}$.
Formally,
\begin{equation}
\hat{\sigma} = \argmin_{\sigma \in \mathfrak{S}_{N}} \sum_i^N  \mathcal{L}_{match}(y_i, \hat{y}_{\sigma(i)}).
\end{equation}

The transformer-based baseline model YOLOPose utilizes a CNN backbone to extract image features and an encoder-decoder architecture followed by feed-forward networks for generating a set of object predictions.
Since the transformer architecture is permutation invariant, the image features are supplemented with positional encoding~\citep{vaswani2017attention}. The encoder performs self-attention between the image features. 
The decoder performs cross-attention between the encoder output and the learned embeddings known as \emph{object queries}. Object queries are initialized randomly at the beginning of the training and are learned along with the training objective.
While performing cross-attention, the encoder output embeddings act as the \emph{key} and the \emph{value}, whereas the object queries act as the \emph{query}. 
The resulting object embeddings are processed in parallel using multi-layer perceptrons (MLPs) to generate object predictions.
During inference, the object queries are fixed. While YOLOPose achieves impressive results, it relies solely on the attention mechanism to learn joint object detection and pose estimation.
Thus, it does not benefit from incorporating inductive biases into its architecture. To this end, we discuss various design strategies for imbuing inductive biases into the YOLOPose model in the following sections.

\subsection{Local Hierarchical Shifting Window Attention}
\label{sec:LHSW}

The encoder module in YOLOPose utilizes self-attention between the image features extracted by a CNN backbone.
This enables aggregating information from all spatial locations in the encoder module.
In terms of the design, one of the main shortcomings of YOLOPose is that the backbone module is exclusively CNN-based and the encoder module is exclusively attention-based. 
Thus, the backbone module does not benefit from the self-attention mechanism. \textit{Vision Transformer models} (ViT)~\citep{dosovitskiy2020image} addressed this issue by designing
a backbone model based exclusively on the transformer architecture. 
\Citet{raghu2021vision} noted that in ViT, the similarity of the lower layer features and the higher layer features is stronger than in the case of CNN-based ResNet model. Based on this observation, the authors concluded that the self-attention mechanism along with the skip connections
enables lower layers in the ViT model to learn global features. However, ViT suffers from two major limitations: 

\begin{enumerate}
  \item feature maps used in the model are low resolution and
  \item complexity of the attention mechanism increases quadratically.
\end{enumerate}
These factors limit the suitability of ViT as a backbone model.
To address this issue, \citet{liu2021Swin} proposed to incorporate hierarchical processing and local sliding window attention in the design of the transformer model.
The pixels are divided into crops. All pixels belonging to an image crop are projected linearly to features of dimension $d$.
The hierarchical processing of the features is shown in~\cref{fig:swin_hierarchical}.
Unlike ViT, in which all the image patches in any particular layer interact with all other patches, the interaction is restricted to a local window (shown in~\cref{fig:swin_windows}).
The layers of the model are grouped into blocks. The size of the attention window and the number of feature maps increases progressively higher in the hierarchy.
The attention window is shifted between the layers in the same block. This ensures the global interaction of the features within each block.
The hierarchical processing of features and limiting the attention to a local window enables linearly scaling computational complexity.
We refer to our implementation of the pose estimation model with local hierarchical shifting window attention as Model-A.


\subsection{Deformable Multi-resolution Attention}
\label{sec:MR-DMHA}
The attention mechanism offers a simple yet effective mechanism to model long-range dependencies in input tokens. Despite the advantages the attention mechanism offers for computer vision tasks, the computational cost of MHA, especially for high-resolution images, remains high.
To address this issue, \citet{zhu2021deformable} introduced \textit{deformable multi-head attention} (DMHA).
In~\cref{fig:model}, we depict the DMHA model for pose estimation.
For a query token $z$, instead of computing attention over all the key tokens $\Omega_x$, DMHA computes attention over a 
small set of 2D reference points $p\in\Omega_p$. The reference points are allowed to deform and the deformation is
learned from the input tokens. Formally, MHA, introduced in~\cref{eqn:mha}, is extended as,
\begin{multline}
\text {DMHA}(z_q, p_q, x) = \\ \sum_{m=1}^{M} W_m \left[  \sum_{k=1}^{K} A_{mqk} \cdot W^{'}_mx(p_q + \Delta p_{mqk})) \right], 
\end{multline}
where $k$ represents the sampled key tokens and $\Delta p_{mqk}$ denotes deformation offset.
Bilinear interpolation enables fractional offsets.
Furthermore, we incorporate multi-resolution feature processing (MR-DMHA) into the deformable attention:
\begin{multline}
\text{MR-DMHA} (z_q, p_q, {\{x\}}^L_{l=1})= \\ \sum_{m=1}^{M} W_m \left[  \sum_{l=1}^{L} \sum_{k=1}^{K} A_{mqk} \cdot W^{'}_mx^{l}(\phi(\hat{p}_q) + \Delta p_{mqk}) \right],
\end{multline}
where $l$ represents the feature level, and the function $\phi(\hat{p}_q)$ re-scales the pixel coordinates corresponding to the feature level.
We refer to our multi-object pose estimation model based on  MR-DMHA as Model-B.

\subsection{Early Fusion of Object Queries}
\label{sec:fusion}
\begin{figure}
  \centering
  \resizebox{.8\linewidth}{!}{\input{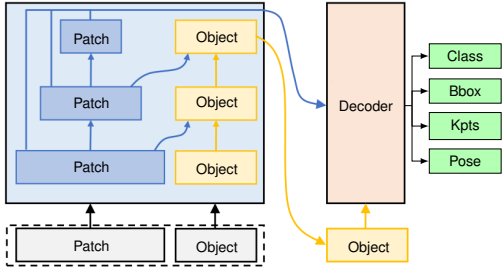}}
  \vspace*{-3mm}
    \caption{Early fusion of object queries. The patch embeddings and the object queries interact within the encoder module at all layers.
    Thus, not only the high-level features interact with the object queries but the features of all resolutions.}
    \label{fig:fused_model}
\end{figure}

While the MR-DMHA enables efficient attention with linear complexity, the encoder module interacts with the object queries only at the final encoder layer.
This results in the encoder module learning generic features in the earlier layers and only the final layer learning features relevant to object prediction.
Enabling encoder-object query interaction at the early layers helps the encoder to focus on features more relevant to the object predictions. 
Following \citet{song2022vidt}, we introduce cross-attention between object queries and patch embeddings early in the encoder module.
In each encoder layer, self-attention is performed between the patch embeddings and additionally, cross-attention is performed between object queries and patch embeddings of the previous layer, as shown in~\cref{fig:fused_model}.
In the decoder module, cross-attention is performed between the aggregated patch embeddings from the encoder and the object query to generate object embeddings.
The object embeddings are processed by the FFNs to generate object predictions. We call this variant Model-C.

\subsection{Query Aggregation}
\label{q-agg}
\begin{figure}
    \centering
    \resizebox{.8\linewidth}{!}{\input{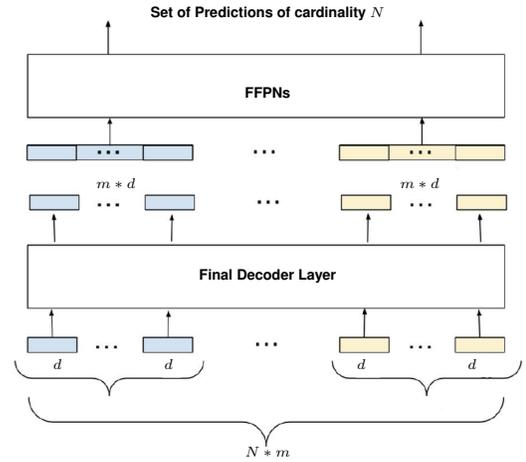}}
      \vspace*{-7mm}
      \caption{Query Aggregation. In contrast to the standard approach of using the same number of object queries as the cardinality of the set $N$,
      in the query aggregation approach, we set the number of object queries to $N \times m$. The query aggregation factor $m$ is a hyperparameter.
      $N \times m$ decoder output embeddings of dimension $d$ are concatenated to form $N$ object embeddings of dimension $d \times m$ that are processed by the FFNs to generate $N$ object predictions.}
      \label{fig:q_agg}
\end{figure}

The learned object queries play a crucial role in DETR-like architectures.
Despite their importance, the object queries are not fully understood.
One of the major reasons behind this lack of understanding
is the fact that neither the architecture itself nor the loss function contains any mechanism to bind the object queries 
specifically to object classes or locations. \citet{zhang2022WhatAE} observed that using more object queries improves model accuracy.
In our models, the number of object queries equals the cardinality of the set we predict.
Thus, increasing the number of object queries also increases the computation cost of bipartite matching and thus, the overall training time.
We propose a novel approach for increasing the number of object queries while keeping the computational cost low.
We call this method \textit{query aggregation} (shown in \cref{fig:q_agg}).
To generate a set of predictions of cardinality $N$, the standard approach uses $N$ object queries of dimension $d$.
In contrast to the standard approach, in the query aggregation approach, we set the number of object queries to $N \times m$, where the aggregation factor $m$ is a hyperparameter.
After the last decoder layer, we concatenate each set of $m$ embeddings to generate one object embedding. Thus, from $N \times m$ output embeddings we generate $N$ object embeddings.
The FFNs process the $N$ object embeddings to generate a set of object predictions. By decoupling the number of decoder output embeddings and the number of object embeddings,
we enable a larger number of embeddings in the decoder layer without increasing the cardinality of the predicted set.

\subsection{Refinement of the Object Predictions}
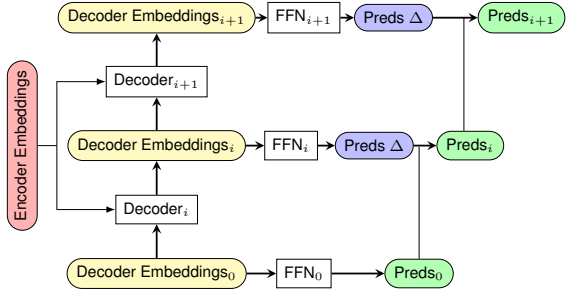
\begin{figure}
  \centering
  \resizebox{.85\linewidth}{!}{\begin{tikzpicture}[font=\scriptsize\sffamily,
arrow/.style = {thick,-stealth}
]
\node(enc_emb)  at (0.1,  1.5) [draw, rounded rectangle, align=center, rotate=90,fill=red!30 ]{Encoder Embeddings};

\node(dec_emb3)  at (2.2,  3.5) [draw, rounded rectangle, align=center,fill=yellow!30]{Decoder Embeddings$_{i+1}$};
\node(dec_emb2)  at (2.2,  1.5) [draw, rounded rectangle, align=center,fill=yellow!30]{Decoder Embeddings$_{i}$};
\node(dec_emb1)  at (2.2,  -0.5) [draw, rounded rectangle, align=center,fill=yellow!30]{Decoder Embeddings$_{0}$};

\node(Dec2)  at (2.2,  2.5) [draw, rectangle, align=center] {Decoder$_{i+1}$};
\node(Dec1)  at (2.2,  0.5) [draw, rectangle, align=center]{Decoder$_{i}$};

\draw [arrow] (dec_emb1.north)  -- (Dec1.south);
\draw [arrow] (Dec1.north)  -- (dec_emb2.south);
\draw [arrow] (dec_emb2.north)  -- (Dec2.south);
\draw [arrow] (Dec2.north)  -- (dec_emb3.south);

\draw [>=latex,->] ($(enc_emb.south)+(0.3, 0)$)  |- (Dec1.west);
\draw [>=latex,->] ($(enc_emb.south)+(0.3, 0)$)  |- (Dec2.west);

\draw [] ($(enc_emb.south)$)  |- ($(enc_emb.south)+(0.3, 0)$);

\node(ffn1)  at (4.5,  -0.5) [draw, rectangle, align=center,]{FFN$_{0}$};
\node(ffn2)  at (4.3,  1.5) [draw, rectangle, align=center,]{FFN$_{i}$};
\node(ffn3)  at (4.5,  3.5) [draw, rectangle, align=center,]{FFN$_{i+1}$};

\node(delta3)  at (5.9,  3.5) [draw, rounded rectangle, align=center,fill=blue!25]{Preds $\Delta$};
\node(delta2)  at (5.6,  1.5) [draw, rounded rectangle, align=center,fill=blue!25]{Preds $\Delta$};

\node(preds3)  at (7.9,  3.5) [draw, rounded rectangle, align=center,fill=green!30]{Preds$_{i+1}$};
\node(preds2)  at (7.1,  1.5) [draw, rounded rectangle, align=center,fill=green!30]{Preds$_{i}$};
\node(preds1)  at (6.3,  -0.5) [draw, rounded rectangle, align=center,fill=green!30]{Preds$_{0}$};

\draw [arrow] (dec_emb1.east)  -- (ffn1.west);
\draw [arrow] (dec_emb2.east)  -- (ffn2.west);
\draw [arrow] (dec_emb3.east)  -- (ffn3.west);

\draw [arrow] (ffn3.east)  -- (delta3.west);
\draw [arrow] (ffn2.east)  -- (delta2.west);
\draw [arrow] (ffn1.east)  -- (preds1.west);

\draw [arrow] (delta3.east)  -- (preds3.west);
\draw [arrow] (delta2.east)  -- (preds2.west);

\draw [] ($(preds1.north)$)  |- ($(preds1.north)+(0, 1.75)$);
\draw [] ($(preds2.north)-(0.1, 0)$)  |- ($(preds2.north)+(-0.1, 1.77)$);
\end{tikzpicture}}
    \caption{Prediction refinement. At each decoder layer, only a small $\Delta$ is predicted on top of the previous predictions enabling the refinement of predictions through the decoder layers.}
    \label{fig:refinement}
\end{figure}

The decoder module in the standard architecture consists of six decoder layers. The output embeddings of each of the decoder layer serves an input for the subsequent decoder layer.
The output embeddings of the final decoder layer are processed by the FFNs to generate object predictions.
Instead of generating object predictions directly once, generating an initial prediction and refining it to generate the final object predictions allows the model to iteratively improve the initial predictions as well as the overall accuracy.
The design of decoder layers naturally suits refinement. 
In~\cref{fig:refinement}, we show the refinement of the object predictions in the decoder module.
Each decoder layer contains its own independent set of FFNs.
The first decoder layer performs cross-attention between the encoder embeddings and the object queries to generate decoder embeddings that are processed by a corresponding set of FFNs to generate object predictions.
The FFNs of subsequent decoder layers generate only a small $\Delta$ to refine the predictions made by the previous layer.
This allows for the model to iteratively refine the initial predictions and generate more accurate final predictions.

\subsection{Loss Function}
Our model is trained to minimize the Hungarian loss between the predicted and the ground-truth sets.
The Hungarian loss is the dissimilarity measure between the matched predicted and the ground-truth pairs. 
To establish the matching pairs, we employ the differentiable bipartite matching algorithm~\citep{stewart2016end, carion2020end}.

The Hungarian loss we use consists of four components: 
bounding box, class probability, translation, and key points.
However, the matching cost we employ comprises only the
bounding box and the class probability components.

Given the matching ground-truth and predicted sets $\mathcal{Y}$ and $\hat{\mathcal{Y}_\sigma}$, respectively, the Hungarian loss is computed as:
{


\begin{multline}\label{hungarian_loss}
\mathcal{L}_{Hungarian}(
   \mathcal{Y}, \hat{\mathcal{Y}_\sigma}) = \sum_i^N \mathcal{L}_{class}+ 
   \mathbbm{1}_{c_i\neq \text{\O}} 
   \Bigl[  \mathcal{L}_{box}(b_i, \hat{b}_{\hat{\sigma}(i)})  \\
                       +  \lambda_{kp} \mathcal{L}_{kp}(k_i, \hat{k}_{\hat{\sigma}(i)})
                       +  \lambda_{pose} \mathcal{L}_{pose}(R_i, t_i, \hat{R}_{\hat{\sigma}(i)}, \hat{t}_{\hat{\sigma}(i)})\Bigr].
\end{multline}
}
The individual components of the Hungarian loss are the following.
\subsubsection{Class Probability Loss}

 \begin{equation}\label{eqn:prob_loss}
 \mathcal{L}_{class}(c_i, \hat{p}_{\sigma(i)}) = -\text{log }\hat{p}_{\hat{\sigma}(i)}(c_i),
 \end{equation}

The standard negative log-likelihood is used to measure the dissimilarity of the predicted class probability scores $\hat{p}_{\sigma}$
and the ground-truth class labels $c$, represented as one-hot encoding.
The images in the YCB-Video dataset~\citep{xiang2017posecnn} comprise between three and eleven objects per frame.
In our model, the cardinality of the set $N$ is set to 20.
Thus, padding the ground-truth set with \text{\O} class creates a heavy class imbalance between the \text{\O} class
and the rest of the classes.
To counter the class imbalance, we weigh the loss for the \text{\O} class with a factor of 0.1.

\subsubsection{Bounding Box Loss}
A weighted combination of the Generalized IoU (GIoU)~\citep{rezatofighi2019generalized} 
and $\ell_1$-loss is employed as the measure of dissimilarity between the predicted and the ground-truth bounding boxes,
$b_i$ and $b_{\sigma(i)}$, respectively:
{
 \begin{equation}\label{eqn:box_loss}
 \mathcal{L}_{box}(b_i, \hat{b}_{\sigma(i)}) = \alpha \mathcal{L}_{iou}(b_i, \hat{b}_{\sigma(i)}) + \beta || b_i - \hat{b}_{\sigma(i)} ||,
 \end{equation}
}
 where $\alpha$ and $\beta$ are hyperparameters.

\subsubsection{Keypoint Loss}
\label{sec:kploss}
The keypoint loss is a combination of $\ell_1$ loss and cross-ratio consistency loss:
{
\begin{equation}\label{eqn:keypoints_loss}
    \mathcal{L}_{kp}(K_i, \hat{K}_{\hat{\sigma}(i)}) = \gamma ||K_i - \hat{K}_{\hat{\sigma}(i)}||_1 + \delta \mathcal{L_{CR}},
\end{equation}
}
where $K_i$ and $\hat{K}_{\hat{\sigma}(i)}$ are ground truth and the predicted keypoint coordinates, respectively, $\mathcal{L_{CR}}$ is the  Smooth$\ell_1$ loss between the ground truth and predicted cross-ratio, and
$\gamma$ \& $\delta$ are weighting factors.


\subsubsection{Pose Loss}
We compute the dissimilarity between the ground-truth pose parameters $R_i$ and $\mathbf{t}_i$,
and the predicted pose parameters $\hat{R}_{\sigma(i)}$ and $\hat{\mathbf{t}}_{\sigma(i)}$
using the disentangled pose loss:

{
\begin{multline}\label{eqn:ploss}
\mathcal{L}_{pose}(R_i, \mathbf{t}_i, \hat{R}_{\sigma(i)}, \hat{\mathbf{t}}_{\sigma(i)}) = \mathcal{L}_{rot}(R_i, \hat{R}_{\sigma(i)}) + \\
+|| \mathbf{t}_i - \hat{\mathbf{t}}_{\sigma(i)} ||_1.
\end{multline}
}

We employ PLoss and SLoss~\citep{xiang2017posecnn} for the rotation component, 
and $\ell_1$ loss for the translation component:

{
\begin{equation}\label{eqn:pose_loss}
\mathcal{L}_{rot} = \left\{\begin{array}{ll}
\frac{1}{|\mathcal{M}_i|} \displaystyle\sum_{\text{x}_1 \in \mathcal{M}_i}  \min_{\text{x}_2 \in \mathcal{M}_i}|| R_i\text{x}_1 - \hat{R}_{\sigma(i)} \text{x}_2 ||_1 \text { if symmetric, } \\
\frac{1}{|\mathcal{M}_i|} \displaystyle\sum_{\text{x} \in \mathcal{M}_i} || R_i\text{x} - \hat{R}_{\sigma(i)} \text{x} ||_1 \text { otherwise, } 
\end{array}\right.
\end{equation}
}
where $\mathcal{M}_i$ indicates the subsampled 3D model points.

\section{Experiments}

\subsection{Dataset}
We evaluate the proposed method on the challenging YCB-Video dataset~\citep{xiang2017posecnn}.
The dataset comprises 92 video sequences of the cluttered tabletop scenes.
Each video sequence consists of a randomly selected subset of objects from a total of 21
objects placed in a random configuration. 
The 6D pose annotations and the bounding box annotations were generated using a semi-automatic procedure 
in which the first frame of each video sequence is manually annotated and extrapolated 
for the rest of the frames employing visual odometry techniques. Overall, the dataset consists of 133,827 images.
Out of the 92 video sequences, twelve are used for testing and the remaining ones are used for training and validation.
We also use the COCO dataset for pre-training our model, which consists of 123,287 images and 886,284 bounding box annotations belonging to 80 categories.

\subsection{Metrics}
\label{sec:metric}

We report the standard area under the curve (AUC) ADD and ADD-S metrics~\citep{xiang2017posecnn}, computed using a varying threshold between 10\,cm to 1\,cm.
Given the ground-truth 6D pose annotation with rotation and translation components $R$ and $\mathbf{t}$, and the predicted rotation and translation components 
$\hat{R}$ and $\hat{\mathbf{t}}$, the ADD metric is the average $\ell_2$ distance between the subsampled mesh points $\mathcal{M}$ in the ground truth and the predicted pose,
whereas the symmetry-aware ADD-S metric is the average $\ell_2$ distance between the closest subsampled mesh points $\mathcal{M}$ in the ground-truth and predicted pose.




The ADD(-S) metric corresponds to ADD-S for symmetric objects and ADD for non-symmetric objects.

\begin{figure}
    \centering
    \resizebox{.8\linewidth}{!}{\begin{tikzpicture}[font=\scriptsize\sffamily,]
   \begin{axis}[
     ymin=65, ymax=95,
     font=\scriptsize,
     tickwidth         = 0pt,
     enlarge x limits  = 0.2,
     xtick={1,2,3,4},
     xtick style={draw=none},
     symbolic x coords = {1,2,3,4},
     nodes near coords,
   legend style={at={(0.00,1.08)},anchor=west},
   legend columns=3,
legend style={/tikz/every even column/.append style={column sep=0.2cm}},
axis y discontinuity=parallel,
    ylabel={Pose Estimation Accuracy },
    xlabel={\# Query Aggregation Factor},
    width=\linewidth,
    height=.7\linewidth,
    nodes near coords style={anchor=south east,inner sep=0,shift={(4pt, 5pt)},font=\scriptsize},
    ylabel near ticks, ylabel shift={-5pt},
   ]

\addplot coordinates { 
 (1,    83.3)
 (2,    84.2)
 (3,    84.6)
 (4,    82.4)
};

\addplot coordinates { 
 (1,    91.2)
 (2,    91.5)
 (3,    91.7)
 (4,    89.9)
};

\addplot coordinates { 
 (1,    69.0)
 (2,    69.0)
 (3,    69.8)
 (4,    66.1)
};

\legend{AUC of ADD(-S), AUC of ADD-S, ADD(-S)}

\end{axis}
\end{tikzpicture}}\vspace*{-2mm}
      \caption{Results from query aggregation experiment.}
      \label{fig:q_agg_result}
\end{figure}
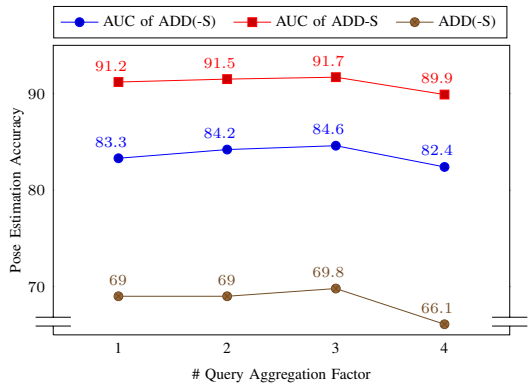


\begin{table}
  \centering
  \begin{threeparttable}
    \caption{Effect of pretraining on the COCO dataset for Object Detection.}
    \label{tab:coco_exp}
    \footnotesize
    {
    \begin{tabular}{l|c|c|c}
    \toprule
 \thead{Method} &\thead{AUC of \\ ADD-S}& \thead{AUC of \\ ADD(-S)} & \thead{ADD(-S)}  \\
  \midrule
  
  Without pretraining  & 71.0  & 82.7 & 42.5 \\
  With pretraining  & \textbf{81.4}  & \textbf{90.1} & \textbf{61.1} \\
  \bottomrule 
\end{tabular}
}
\end{threeparttable}
\end{table}

\subsection{Implementation Details}
\label{sec:imp_dtl}
We implement our models using the PyTorch~\citep{NEURIPS2019_9015} library.
Our models are trained using NVIDIA A100 GPUs with 40 GBs of memory. The size of the input images is \mbox{640$\times$480}. 
The models are trained for 200 epochs with a batch size of 32 using AdamW optimizer with an initial learning rate of 10$^{-4}$.
During training, we supplement the training images with the synthetic images provided with the dataset.
The hyperparameters $\alpha$, $\beta$, $\gamma$, and $\delta$ in~\cref{eqn:box_loss,eqn:keypoints_loss} are set to 2, 5, 10, and 1, respectively.

\begin{table*}
\scriptsize
  \centering
	\begin{threeparttable}
	\caption{Results on the YCB-Video dataset~\citep{xiang2017posecnn}.}
  \label{tab:ycbv-details}
 
	\begin{tabular}{l|c|c|c|c|M{1cm}|c|c|c|c|M{1cm}}
	\midrule
  Metric & \multicolumn{5}{c|}{AUC of ADD-S} & \multicolumn{5}{c}{AUC of ADD(-S)} \\
  \midrule
	Method &
  \thead{GDR-Net \\ \citep{Wang_2021_GDRN}} &  
	\thead{DeepIM \\ \citep{li2018deepim}} &
	\thead{YOLOPose \\ \citep{amini2022yolopose}} &
  \thead{YOLOPose  \\ V2 \citep{periyasamy2023yolopose}} &
	\thead{Ours} &
	\thead{GDR-Net \\ \citep{Wang_2021_GDRN}} &  
	\thead{DeepIM \\ \citep{li2018deepim}} &
	\thead{YOLOPose \\ \citep{amini2022yolopose}} &
  \thead{YOLOPose  \\ V2 \citep{periyasamy2023yolopose}} &
	\thead{Ours} \\
	\midrule
	master\_chef\_can        &\textbf{96.6} & 93.1         & 91.3         & 91.7           & 90.3        & 71.1            & 71.2           & 64.0             & \textbf{71.3} & 66.7          \\
	cracker\_box             & 84.9         & 91.0         & 86.8         &  92.0          &\textbf{92.3}& 63.5            & 83.6           & 77.9             & 83.3          & \textbf{86.0} \\
	sugar\_box               &\textbf{98.3} & 96.2         & 92.6         & 91.5           &94.4         & 93.2            & \textbf{94.1}  & 87.3             & 83.6          & 89.1          \\
	tomato\_soup\_can        &\textbf{96.1} & 92.4         & 90.5         & 87.8           & 89.2        & \textbf{88.9}   & 86.1           & 77.8             & 72.9          & 76.3          \\
	mustard\_bottle          &\textbf{99.5} & 95.1         & 93.6         & 96.7           & 96.5        & \textbf{93.8}   & 91.5           & 87.9             & 93.4          & 93.3          \\
	tuna\_fish\_can          & 95.1         &\textbf{96.1} & 94.3         & 94.9           & 94.5        & 85.1            & \textbf{87.7}  & 74.4             & 70.5          & 67.4          \\
	pudding\_box             & 94.8         & 90.7         & 92.3         & 92.6           &\textbf{95.5}& 86.5            & 82.7           & 87.9             & 87.0          & \textbf{91.9} \\
	gelatin\_box             & 95.3         & 94.3         & 90.1         & 92.2           &\textbf{95.4}& 88.5            & \textbf{91.9}  & 83.4             & 85.7          & 91.8          \\
	potted\_meat\_can        & 82.9         & 86.4         & 85.8         & 85.0           &\textbf{88.9}& 72.9            & 76.2           & \textbf{76.7}    & 71.4          & 76.4          \\
	banana                   &\textbf{96.0} & 91.3         & 90.0         & 95.8           & 95.4        & 85.2            & 81.2           & 88.2             & 90.0          & \textbf{91.0} \\
	pitcher\_base            &\textbf{98.8} & 94.6         & 93.6         & 95.2           & 94.9        & \textbf{94.3}   & 90.1           & 88.5             & 90.8          & 89.9          \\
	bleach\_cleanser         &\textbf{94.4} & 90.3         & 85.3         & 83.1           & 87.3        & 80.5            & \textbf{81.2}  & 73.0             & 70.8          & 73.9          \\
	bowl$^*$                 & 84.0         & 81.4         & 92.3         & \textbf{93.4}  & 91.9        & 84.0            & 81.4           & \textbf{92.3}    & \textbf{93.4} & 91.9          \\
	mug                      &\textbf{96.9} & 91.3         & 84.9         & 95.5           & 95.5        & 87.6            & 81.4           & 69.6             & \textbf{90.0} & 89.3          \\
	power\_drill             & 91.9         & 92.3         & 92.6         & 92.5           &\textbf{94.6}& 78.7            & 85.5           & 86.1             & 85.2          & \textbf{88.9} \\
	wood\_block$^*$          & 77.3         & 81.9         & 84.3         & \textbf{93.0}  &\textbf{93.0}& 77.3            & 81.9           & 84.3             & \textbf{93.0} & \textbf{93.0} \\
	scissors                 & 68.4         & 75.4         &\textbf{93.3} & 80.9           & 89.5        & 43.7            & 60.9           & \textbf{87.0}    & 71.2          & 76.2          \\
	large\_marker            &\textbf{87.4} & 86.2         & 84.9         & 85.2           & 84.5        & 76.2            & 75.6           & 76.6             & 77.0          & \textbf{77.4} \\
	large\_clamp$^*$         & 69.3         & 74.3         & 92.0         & \textbf{94.7}  &\textbf{94.2}& 69.3            & 74.3           & 92.0             & \textbf{94.7} &  94.2         \\
	extra\_large\_clamp$^*$  & 73.6         & 73.3         &\textbf{88.9} & 80.7           & 79.2        & 73.6            & 73.3           & \textbf{88.9}    & 80.7          & 79.2          \\
	foam\_brick$^*$          & 90.4         & 81.9         & 90.7         & 93.8           &\textbf{95.0}& 90.4            & 81.9           & 90.7             & 93.8          & \textbf{95.0} \\
	\midrule 
	Mean                     & 89.1         &88.1          & 90.1         & 91.2           &\textbf{92.0}& 80.2            & 81.9           & 82.6             & 83.3          & \textbf{84.7} \\
	\bottomrule
	\end{tabular}
  The best results are shown in bold.
\end{threeparttable}

	\end{table*}

\begin{figure}
        \centering
        \newlength{\imgres}
        \setlength{\imgres}{0.16\textwidth}
        \setlength{\tabcolsep}{0.01cm}
        \begin{tabular}{ccc}
            
         \includegraphics[width=\imgres]{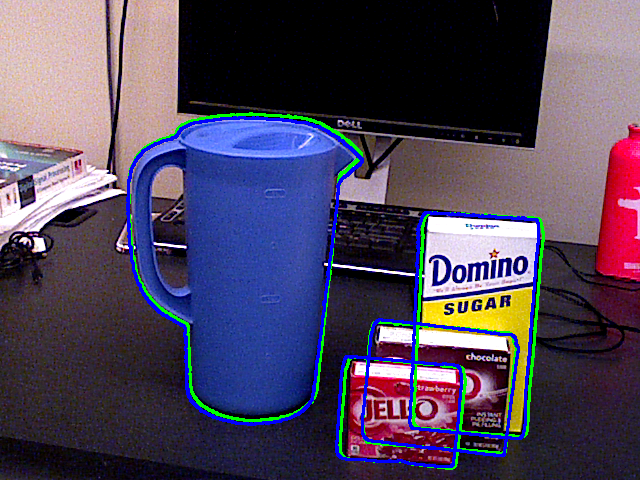} &
         \includegraphics[width=\imgres]{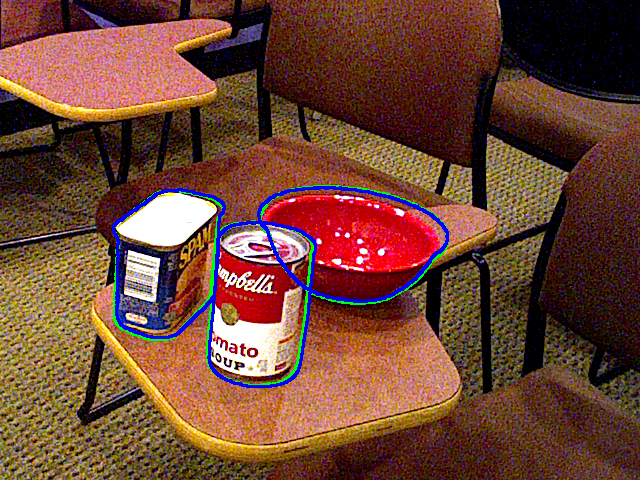} &
         \includegraphics[width=\imgres]{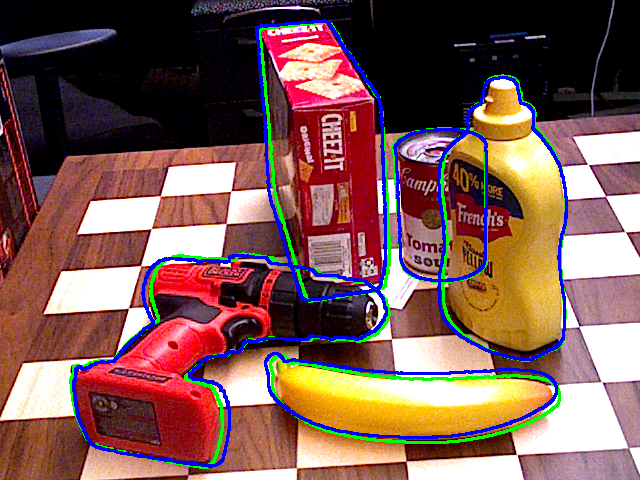} \\
         \includegraphics[width=\imgres]{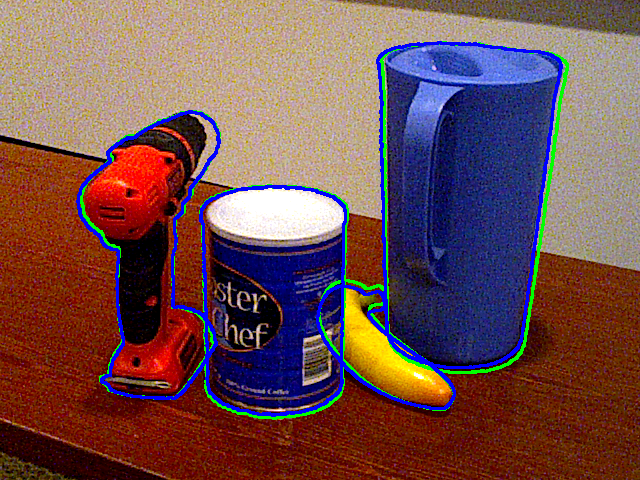} &
         \includegraphics[width=\imgres]{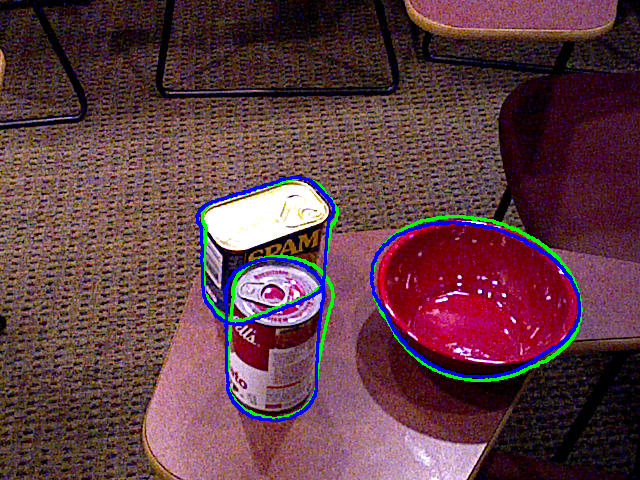} &
         \includegraphics[width=\imgres]{figures/viz/good/129/000052_000527.png} \\

        \end{tabular}
        \caption{Qualitative results on YCB-Video dataset~\citep{xiang2017posecnn}. 
        Ground truth and predicted object poses are visualized as object contours in green and blue colors, respectively.
        }
        \label{fig:result}
\end{figure}

\begin{figure}
	\centering
	\newlength{\imgF}
	\setlength{\imgF}{3.cm}
	\setlength{\tabcolsep}{0.009cm}
	{\footnotesize
	\begin{tabular}{cc}
	 (a)\,\includegraphics[width=\imgF]{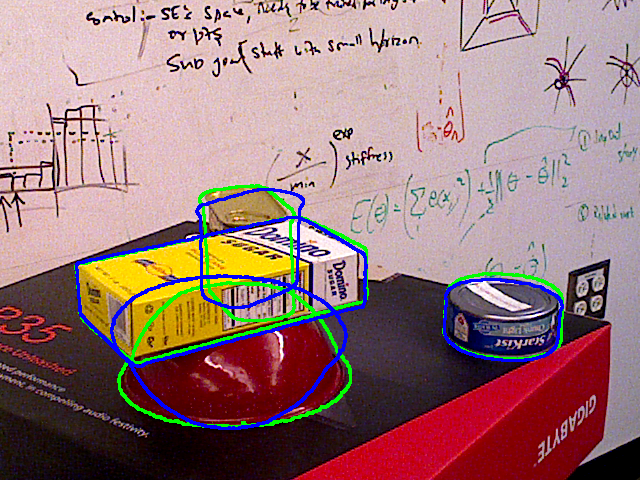} &
	 ~(b)\,\includegraphics[width=\imgF]{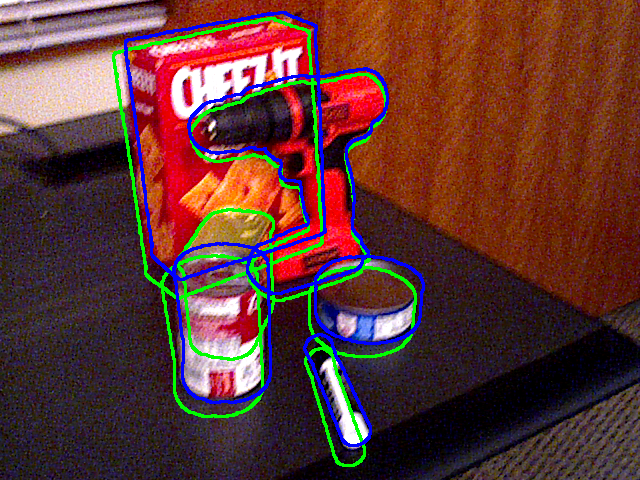}\vspace*{1mm} \\
	 (c)\,\includegraphics[width=\imgF]{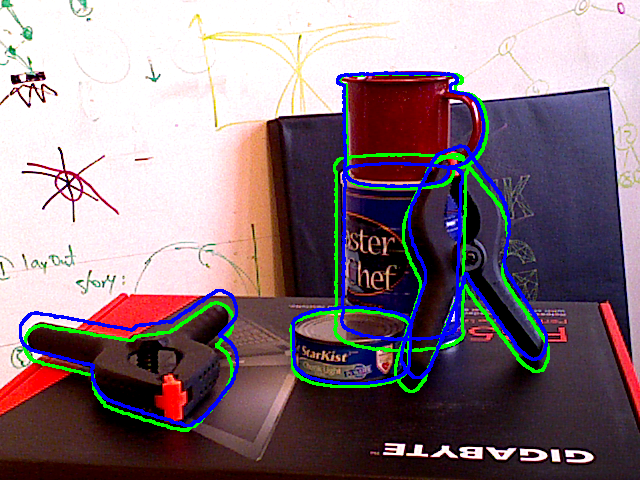} &
	 ~(d)\,\includegraphics[width=\imgF]{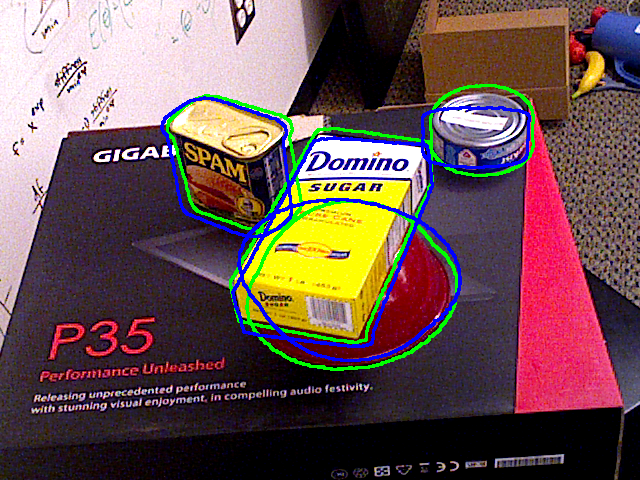} \\
	\end{tabular}
	}
  \caption{Typical failure cases.
	Ground truth and predicted object poses are visualized as object contours in green and blue colors, respectively.
  Occlusion negatively influences the accuracy of our model.
	}
  \label{fig:fail_case}
\end{figure}

\begin{table}
  \scriptsize
  \centering
  \begin{threeparttable}
    \caption{Results from Ablation Study.}
    \label{tab:ablation}
        {
    \begin{tabular}{l|c|c|c|c}
    \toprule
 \thead{Method} &\thead{AUC of \\ ADD-S}& \thead{AUC of \\ ADD(-S)} & \thead{Params \\$ \times 10^6 $ } & \thead{fps} \\
  \midrule
  YOLOPose~\citep{amini2022yolopose}  & 90.1 & 82.6 & \textbf{48.6} & \textbf{41.8} \\
  YOLOPose V2~\citep{periyasamy2023yolopose}  & 91.2 & 83.3 & 53.2 & 39.1 \\
  CosyPose~\citep{labbe2020}  & 89.8 & 84.5 &  - & 2.5 \\
  \midrule
  Model-A  & 90.1 & 81.4  & 57 & 39.5\\
  Model-B 16 def pts  & \textbf{92.0} & \textbf{84.7} & 55.2 & 25.9 \\
  Model-B 6 def pts  & 81.9 & 89.3 & 51.7 & 28.1 \\
  Model-B 6 def pts + refine  & 90.2 & 83.0 & 87.1 & 25.6 \\
  Model-C  & 90.3  & 82.1 & 48.7 & 34.8\\
  \bottomrule 
\end{tabular}
}
\end{threeparttable}
\begin{tablenotes}
  \scriptsize
\item[] Model-A: Local hierarchical attention-based model discussed in~\cref{sec:LHSW}.
\item[] Model-B: Model based on multi-resolution deformable multi-head attention discussed in~\cref{sec:MR-DMHA}.
\item[] Model-C: Model utilizing early fusion discussed in~\cref{sec:fusion}.
\item[] def pts: Number of deformable points. 
\end{tablenotes}
\vspace{-3mm}
\end{table}

\subsection{Results}
In~\cref{fig:result}, we show pose estimation for exemplar scenes; and in~\cref{tab:ycbv-details}, we report the quantitative comparison of our model with the state-of-the-art RGB pose estimation models.
The best-performing variant of our model is based on deformable multi-resolution attention (Model-B) discussed in~\cref{sec:MR-DMHA} utilizing 16 deformable points. 
Our model achieves an impressive AUC of ADD-S score of 92.0 and AUC of ADD(-S) score of 84.7.
Among the object categories, our model performs worse for \emph{extra-large clamp} and \emph{scissors}--both exhibit challenging geometry.
In~\cref{fig:fail_case}, we show typical failure cases of our model. \emph{bowl}, for example, is often predicted upside down. Occlusion still remains a big challenge for our model.
In~\cref{tab:ablation}, we compare the accuracy of the different models we discussed in~\cref{sec:method}. 

In comparison to \mbox{Model-B}, \mbox{Model-A} performs a little worse.
This demonstrates that despite the careful design employing shifted window attention, the model suffers from 
inefficiencies in global dependency modeling. Moreover, \citet{zhou2022understanding} noted that the models based on local hierarchical shifting windows suffer from a lack of robustness.
Although \mbox{Model-C}, based on the early fusion of object queries, performed better than \mbox{Model-A}, it did not match the performance of Model-B.
However, in terms of the inference speed, \mbox{Model-A} performs better than other models---except for the baseline YOLOPose model. \mbox{Model-B} is the slowest. This is caused by accessing random memory locations in the deformable attention computation.

\subsection{Ablation Study}
\label{sec:ablation}

\subsubsection*{Query Aggregation}
In \cref{fig:q_agg_result}, we present the results of training our model with different query aggregation factors.
The performance of the model increases with the query aggregation factor and reaches the highest accuracy for factor 3 but drops for factor 4.

\subsubsection*{Need for COCO Pre-training}
Training the vision transformer models for set prediction is harder than training CNN models to perform single object pose prediction due to the usage of bipartite matching to find the matching pairs between the predicted and the ground-truth sets, which results in slower convergence.
Moreover, training data requirements are also much larger for the set predictions task, compared to single object prediction.
We hypothesize that in the initial phase of the training, the model learns to detect multiple objects in the image and only in the later stages the model learns to predict keypoints and the pose parameters.
Although the YCB-Video dataset~\citep{xiang2017posecnn} is considerably larger than the other pose annotation datasets, it is not big enough to train vision transformer models for multi-object pose estimation.
To overcome this limitation, we train our model initially on the COCO dataset for the task of multi-object detection (class probability and bounding box prediction) and then train the model for multi-object pose estimation on the YCB-Video dataset.
In~\cref{tab:coco_exp}, we compare the pose estimation accuracy of models trained using only the YCB-Video dataset~\citep{xiang2017posecnn} and using COCO dataset for pretraining.
The model pretrained using the COCO dataset outperforms the model trained using only the YCB-Video dataset, highlighting the importance of large-scale pretraining for training vision transformers to learn the task of multi-object prediction.

\subsubsection*{Prediction Refinement}
In~\cref{tab:ablation}, we present the results of the prediction refinement experiment.
Refinement boosted the performance of Model-B constructed with six deformable points by 0.9 and 1.1 accuracy points in terms of the \mbox{AUC of ADD-S} and \mbox{AUC of ADD-(S)} metrics, respectively.
However, the improvements come at a cost of an increased number of parameters: 87.1$\times 10^6$ compared to 51.7$\times 10^6$.
Interestingly, for the Model-B constructed using 16 deformable points that achieves an impressive accuracy of \mbox{92 AUC of ADD-S}  and 84.7 \mbox{AUC of ADD-(S)}, the boost in performance is negligible.

\subsection{Limitations}
While formulating multi-object pose estimation as a set prediction problem facilitates the prediction of a varying number of objects in the given image, training the model needs complete set annotations for all objects in the given image. Most of the commonly used pose estimation datasets like Linemod-Occluded~\citep{Linemodoccluded} and Linemod~\citep{hinterstoisser2013model} offer only partial annotations for training images.
Thus, they are unsuitable for training our models.
Acquiring complete pose annotations can be prohibitively expensive in real-world settings. 

\section{Conclusion}
In this paper, we investigated various inductive biases in the design of multi-object pose estimation models, namely,
local hierarchical shifting window attention, deformable multi-resolution attention, and early fusion of object queries.
Moreover, we proposed a query aggregation mechanism to increase the number of object queries without increasing the computational complexity of our model.
The best-performing model based on deformable multi-resolution attention achieves state-of-the-art results on the challenging YCB-Video dataset.

\section{Acknowledgment}
This work has been funded by the German Ministry of Education and Research (BMBF), grant no. 01IS21080,
project ``Learn2Grasp: Learning Human-like Interactive Grasping based on Visual and Haptic Feedback''.

\printbibliography

\end{document}